# Multi-level feature fusion network combining attention mechanisms for polyp segmentation

*Junzhuo Liu [a], Qiaosong Chen [a],\*, Ye Zhang [a], Zhixiang Wang [b], Xin Deng [a], Jin Wang [a]*

[a]Key Laboratory of Data Engineering and Visual Computing, Chongqing University of Posts and Telecommunications, Chongqing 400065,P.R.China.
[b]Department of Radiation Oncology (Maastro), GROW School for Oncology and Reproduction, Maastricht University Medical Centre+, Maastricht, Netherlands.

A B S T R A C T

Clinically, automated polyp segmentation techniques have the potential to significantly improve the efficiency and accuracy of medical diagnosis, thereby reducing the risk of colorectal cancer in patients. Unfortunately, existing methods suffer from two significant weaknesses that can impact the accuracy of segmentation. Firstly, features extracted by encoders are not adequately filtered and utilized. Secondly, semantic conflicts and information redundancy caused by feature fusion are not attended to. To overcome these limitations, we propose a novel approach for polyp segmentation, named MLFF-Net, which leverages multi-level feature fusion and attention mechanisms. Specifically, MLFF-Net comprises three modules: Multi-scale Attention Module (MAM), High-level Feature Enhancement Module (HFEM), and Global Attention Module (GAM). Among these, MAM is used to extract multi-scale information and polyp details from the shallow output of the encoder. In HFEM, the deep features of the encoders complement each other by aggregation. Meanwhile, the attention mechanism redistributes the weight of the aggregated features, weakening the conflicting redundant parts and highlighting the information useful to the task. GAM combines features from the encoder and decoder features, as well as computes global dependencies to prevent receptive field locality. Experimental results on five public datasets show that the proposed method not only can segment multiple types of polyps but also has advantages over current state-of-the-art methods in both accuracy and generalization ability. Code is available at https://github.com/ngfufdrdh/MLFFNet.

## 1. Introduction

The incidence of cancer is increasing worldwide due to the aging of the population and increased risk factors. Cancer is gradually becoming one of the main causes of unnatural death in the population in various countries. According to[1], nearly 10 million people died of cancer in 2020, accounting for about one-sixth of all deaths worldwide. Among all cancers, colorectal cancer is the third most prevalent, after breast cancer and lung cancer. Most colorectal cancers are caused by malignant lesions of benign adenomas. Adenomas begin as colonic polyps and can lead to cancer when they deteriorate[2]. Early polyp screening can significantly reduce the incidence of colorectal cancer. Colonoscopy is considered the gold standard for CRC screening[3]. However, due to the number and location of polyps and the subjective judgment of physicians, about 25% of polyps are missed in clinical cases[4][5]. To aid in clinical diagnosis, improve detection accuracy and reduce patient disease risk, there has been a growing focus on the auto-segmentation of polyps.

Early automatic polyp segmentation algorithms frequently use nonlinear diffusion over rate[6], shape prior-based boundary detection[7], clustering[8][9][10], and so on. However, these traditional polyp segmentation methods are difficult to be extensively applied in clinical situations due to low accuracy, high data specificity, and severe prior knowledge dependence.

Automatic polyp segmentation based on deep learning shows a significant improvement in both accuracy and generalization over traditional methods. Hemin[11] et al. utilized multiple convolutional neural networks[12][13] as feature extractors and integrated a tuned Mask R-CNN[14] as a segmentation network to simultaneously improve the accuracy of polyp segmentation and detection. To recognize and segment polyps, Patrick[15] et al. modified a fully convolutional network[16] (FCN). Lin[17] et al. modified U-Net[18] by fusing global contexts and generating discriminative attention features with residual attention to achieve complementary global and local refinement in polyp segmentation. Overall, the deep learning-based automatic polyp segmentation method reduces the reliance on prior knowledge, circumvents the limitations of hand-designed features, and greatly enhances generalizability. Nevertheless, the current polyp segmentation algorithm is still not easily used widely in the clinic, and the main problems are related to the following two aspects: 1) inadequate feature utilization, and 2) semantic conflicts and information redundancy in the fusion process of features at all levels. The existing studies are mostly based on the encoder-decoder framework. The features of each layer formed by multiple downsampling of the encoder differ at different levels. Specifically, the shallow-level features contain rich spatial detail features and the deep-level features contain strong semantic features. There are also representational differences at different levels in the deep features. However, most studies do not take full advantage of these features. In the process of feature fusion, features of different levels complement and enhance themselves through direct combination. The majority of current studies, however, have not taken into account information conflicts and redundancy bias brought about by the fusing of features with various resolutions, as well as semantic disparities in the fusion process.

\* *Corresponding author.*
  E-mall addresses: chenqs@cqupt.edu.cn (Q. Chen)

To address the above issues, we propose a multi-level feature fusion network combining attention mechanisms, called MLFF-Net. MLFF-Net aims to complement features located at different levels by feature fusion, including multi-level encoder feature fusion and encoder-decoder feature fusion. Attention is used to mitigate conflicting information redundancy during feature fusion and enhance shallow detail information. Specifically, the encoder output is divided into shallow features rich in detailed information and deep features rich in semantic information, which can be considered separately. Multi-scale Attention Module(MAM) extracts multiscale information and details such as texture and boundaries from shallow features. High-level Feature Enhancement Module (HFEM) aggregates the semantic information in deep features from deep to shallow to make multi-level features complementary to each other. The attention mechanism in HFEM refilters the fused features to mitigate the semantic differences and conflicting redundancies from different layers. Instead of directly concatenating the features from the encoder with the decoder features, as most studies do, we design a Global Attention Module(GAM). GAM, which is part of the decoder, avoids the limitations of position sensitivity by fusing same-level decoder features and encoder features to capture global relationships. The contributions of this study are summarized below.

To solve the problems of inadequate feature utilization and conflicting feature fusion in existing studies in the field of polyp segmentation, we propose a novel polyp segmentation network based on multilayer feature fusion and attention mechanisms.

We design three modules to improve the baseline network. MAM is used to capture multi-scale information and minute details in the shallow features of the encoder. HFEM aggregates deep features of the encoder from deep to superficial to more accurately locate polyps. The attention mechanism in HFEM can mitigate the semantic discrepancies and information redundancy caused by feature fusion. To capture global information in decoder features, we propose GAM, which combines same-level encoder and decoder features and computes global dependencies from them.

Extensive experiments on five public polyp datasets have shown that MLFF-Net has excellent learning and generalization capabilities and can accurately segment multiple types of polyps.

## 2. Related work

### 2.1. Polyp Segmentation

Computer-assisted medical expert diagnosis in a clinical context can greatly increase diagnostic effectiveness and accuracy, lowering patient disease risk. Colonic polyp screening, a key tool for early colorectal cancer screening, has attracted a lot of interest from the medical field and the field of artificial intelligence for its automatic segmentation technology. These studies are divided into traditional segmentation methods and deep learning-based segmentation methods.

Traditional segmentation methods are mostly based on the color, texture, and shape of the polyps. Nima[19] et al. proposed a hybrid context-shape approach. It reliably locates polyps by edge classification to remove non-polyp structures, which involves a coarse edge map of the image and contextual information. [20] proposed a texture-based feature approach that uses Student's test to filter the features, and a support vector machine[21] (SVM) was used to identify polyp locations. [22] proposed a color, texture-based feature extraction method for colon images using Principal Component Analysis(PCA)[23] and nearest neighbors to obtain parameters in the color and achromatic domains of the images.

Most deep learning-based polyp segmentation methods rely on large amounts of the training dataset, following the encoder-decoder framework. The encoder is used for feature extraction of the image dataset, and the decoder works on polyp segmentation and detection. ACSNet[24] found that comparative inference of local and global contextual information is beneficial to improve the accuracy of segmentation of polyps of different sizes, designing channel attention mechanisms for adaptive selection and aggregation of the extracted contextual information. Similarly, Krushi[25] et al. designed a semantic feature enhancement module based attention mechanism for filtering important features and attenuating noise. Moreover, encoder features were passed through an adaptive global context module where important and hard fine-grained features were fused with decoder features. SANet[26] designed a color exchange strategy for the color inconsistency issue of the dataset samples. It eliminated the interference caused by color inconsistency due to variations in acquisition conditions, thus allowing the model to focus more on polyps. For the degradation of small polyps during downsampling, a shallow attention module was proposed to preserve small polyp structures by filtering background noise. LODNet[27] divided the polyp segmentation task into two stages. In the first stage, the eight directional derivatives of each pixel in the target region were calculated, and the pixels with larger derivative values constituted the target candidate regions. In the second stage, the candidate region boundaries were refined by fusing semantic features and boundary features to improve the segmentation accuracy of polyp edges. These studies above are mainly based on convolutional neural networks. With the proposal of the transformer[28], there are some studies that apply this novel framework to the field of polyp segmentation. Dong[29] et al. used PVTv2[30] as an encoder, which can learn more powerful, robust representations from samples. For low-level local features and high-level cue features, this study combined graph convolution[31] and non-local modules[32] to inject local features into high-level semantic features. Du[33] et al. suggested a medical image segmentation algorithm based on Swin Transformer[34], which used dense connectivity and local pyramidal attention to aggregate contextual information and reduced background noise interference. This method was not only suitable for polyp segmentation, but also had high applicability for other segmentation tasks.

*2.2. Attention Mechanism*

Models always require a large amount of training data as input in deep learning. The information that is beneficial for model learning only makes up a small portion of these data, which are heavily contaminated with noise and interference information. To solve this problem, attention mechanism has been proposed. The study of attention mechanisms which was first inspired by the process of human cognition has been applied to several fields such as text classification[35][36][37], machine translation[38][39][40][41], target detection[42][43][44][45][46], image generation[47][48], etc. Hu[49] et al. proposed a channel attention mechanism called SENet. SENet constructed a relational vector of feature map channel dimensions through the squeeze-excitation module, which enabled the adaptive readjustment of feature map weights. BAM[50] suggested a bottleneck attention module that acts on the location of the downsampling feature maps. The BAM incorporated weights calculated separately from the channel dimension and the spatial dimension, thus improving the representational capability of the model.

Attention mechanisms are also widely used in polyp segmentation. PraNet[51] introduced a reverse attention mechanism to improve the segmentation accuracy of polyp edges. Reverse attention divided the feature map into foreground, background, and hard-to-judge edge detail parts, and got attention weights by erasing foreground objects. The parallel axis attention in UCUANet[52] avoided the deformation caused by single-axis attention and used element-by-element summation processing to compensate for possible artifacts in the segmentation process. It balanced global dependence and local detail representation, with the ability to segment polyps more accurately.

# 3. Method

*3.1. Overall Architecture*

In this section, we first present the overall architecture of the model and the process from input to output, as shown in Fig 1. Then, each part is described separately, including Multi-scale Attention Module, High-level Feature Enhancement Module, Global Attention Module, and loss function.

MLFF-Net, as shown in Fig.1, which follows the U-Net architecture containing encoder, decoder, and skip connection, implements polyp segmentation in an end-to-end manner. It consists of four key modules: Encoder following the Transformer architecture, Multi-scale Attention Module (MAM), High-level Feature Enhancement Module (HFEM), and Global Attention Module (GAM). Specifically, the Transformer encoder extracts features and long-range relationships in the input samples by multiple downsampling and outputs a set of feature maps suitable for the segmentation task. MAM first extracts the multiscale information from the shallow encoder feature maps, and then removes the background noise from the shallow information to enhance the detailed information representation. These details include mainly textures, fine edges, and tiny structures. HFEM aggregates multi-level deep features of encoders, which are mainly semantic information, and mitigates the semantic conflicts caused by multi-level fusion. GAM aggregates the decoder features and the skip connection from the encoder. The global relationships computed during aggregation are used to realign the decoder feature maps.

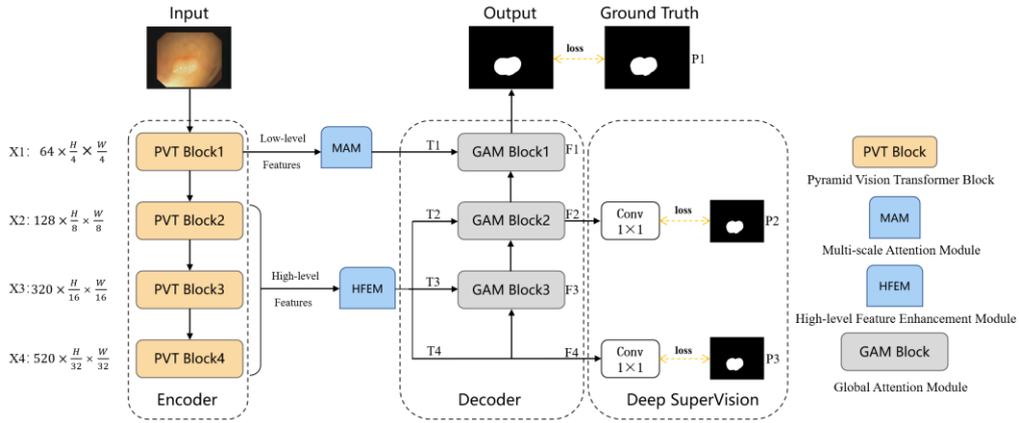

**Fig. 1.** Overall architecture of the proposed Network.

*3.2. Multi-scale Attention Module*

In the polyp segmentation task, the utilization of shallow features can substantially improve the accuracy. These shallow features contain rich information on details such as textures, boundaries, etc. Meanwhile, the multi-scale features deserve to be noticed because the morphology and size of polyps show a strong inconsistency in the dataset. In general, the model needs to preserve the details of tiny polyps in the cluttered background noise, while also balancing the integrity and boundary accuracy of large polyps.

With the above purpose, we design a multi-scale attention module (MAM) working on shallow features. The multiscale part of MAM is used to extract multiscale features, which consists of four convolutional layers with increasing kernel size and inflated convolution. Different branches utilize various convolutional kernel size k ∈ {1, 3, 5, 7}, providing different receptive fields. The convolution kernel is implemented in k×1 and 1×k, which can significantly reduce the number of parameters with less impact on performance. Dilated convolution in the branch further expands the receptive field to provide better discriminability. The feature maps that go through the multiscale module are concatenated in the channel dimension and summed pixel by pixel with $X_1$ as residual connections.

*3.3. High-level Feature Enhancement Module*

The deep features of the encoder contain rich abstract semantic information, which mainly affects the localization of polyps, the class judgment of pixels, etc. The features are gradually abstracted while the encoder continuously downsamples the features. However, the features are still consistent at all levels. In other words, the most beneficial features for segmentation are present at all levels of encoders, only the specific expression dimensions are different. For these reasons, the model uses a deep-to-shallow approach to aggregate semantic features.

*3.4. Global Attention Module*

The role of the decoder is to transform the abstract feature map into a prediction map by progressive upsampling. In most studies, this process mainly consists of convolution and concatenation of skip connection. However, this approach has unavoidable limitations: the convolution kernel captures only the intrinsic correlation of some pixels within the receptive field while lacking the global grasp of the whole feature map. This localization decreases the accuracy of polyp segmentation, as the texture of the polyp is extremely similar to the background, and the borders are severely blurred.

To solve the above problems, we follow the thought of non-local[32] and design a global attention module(GAM). GAM not only captures global information from the current decoder feature map, but also adds the same-level encoder features to the global weight calculation. The motivation for this strategy is that 1) same-level features can complement each other, and 2) encoder features that have undergone MAM and HFEM have less background

\* *Corresponding author.*
E-mail addresses: chenqs@cqupt.edu.cn (Q. Chen)

noise and possess higher utilization values.

*3.5. Loss Function*

To pay more attention to the parts of polyps that are difficult to segment, we use weighted Intersection over Union (IoU) Loss[55] and weighted binary cross entropy (BCE) Loss[55] as the basic loss functions, denoted as $L_{IoU}^w(.)$ and $L_{BCE}^w(.)$. Unlike the standard IoU Loss and BCE Loss, they assign weights to all pixels and focus more on the pixels that have a significant impact on the loss calculation. where $L_{IoU}^w(.)$ cares about the global structure, and $L_{BCE}^w(.)$ focuses on local details. The basic loss function is expressed as Equation (8).

$$L_b(X,Y) = L_{IoU}^w(X,Y) + L_{BCE}^w(X,Y) \tag{8}$$

$X$ is the predicted output of the model and $Y$ is the Ground Truth.

As illustrated in Fig. 1, the model outputs three prediction maps $P_1$, $P_2$, and $P_3$. Therefore, the loss function contains three components, as shown in Equation (9).

$$L_{all} = L_b(P_1, G) + L_b(P_2, G) + 0.5 \times L_b(P_3, G) \tag{9}$$

$G$ is Ground Truth. $P_1$ and $P_2$ are considered the main prediction maps which are participated in the loss calculation and segmentation prediction of the model. $P_3$ is utilized as an auxiliary prediction map to assist in model training.

# 4. Experimental design

*4.1. Dataset*

Five datasets are used to complete the experiments, including Kvasir[56], CVC-ClinicDB[57], CVC-ColonDB[19], ETIS[58], and CVC-300[59]. A portion of the data from Kvasir and CVC-ClinicDB is taken to train the model.

Kvasir: The dataset is manually annotated by clinicians and validated by gastroenterologists. It contains 1000 colonoscopy images and their masks, with image resolutions ranging from 332 × 487 to 1920 × 1072.

CVC-ClinicDB: The dataset consists of 612 image frames extracted from different colonoscopy sequences and segmentation masks annotated by clinical experts. The resolution of the image is 384 × 288.

CVC-ColonDB: It consists of 380 still images. The resolution of the images is 574 × 500.

ETIS: Includes 196 frames of polyps extracted from the colonoscopy video and their corresponding mask composition. The image resolution is 1225 × 966. Each input image contains at least one polyp.

CVC-300: Consists of 60 colonoscopy images with a resolution size of 574 × 500.

*4.2. Evaluation Criteria*

To comprehensively evaluate the proposed method, multiple evaluation criteria are applied, including mean Sørensen–Dice coefficient ($mDic$), mean Intersection of Union ($mIoU$), weighted F-measure($F_\beta^\omega$)[60], S-measure($S_\alpha$)[61], mean E-measure($mE_\xi$)[62][63], max E-measure ($maxE_\xi$)[62][63], mean absolute error($MAE$)[64].

Both $mDic$ and $mIoU$ measure the similarity between prediction and ground truth. $F_\beta^\omega$ integrates recall and precision to overcome the unfair measurement caused by interpolation defects and dependency defects. $S_\alpha$ assesses the structural similarity between the predicted and ground truth. $E_\xi$ evaluates the segmentation results from both pixel and image levels. $MAE$ indicates the mean absolute error between the predicted and ground truth.

*4.3. Implementation Details*

# 5. Results and discussion

In this section, we compared our method with state-of-the-art methods. Then, we discuss the improvement of the prediction results by each module using ablation experiments. Finally, the limitations of the proposed method are illustrated.

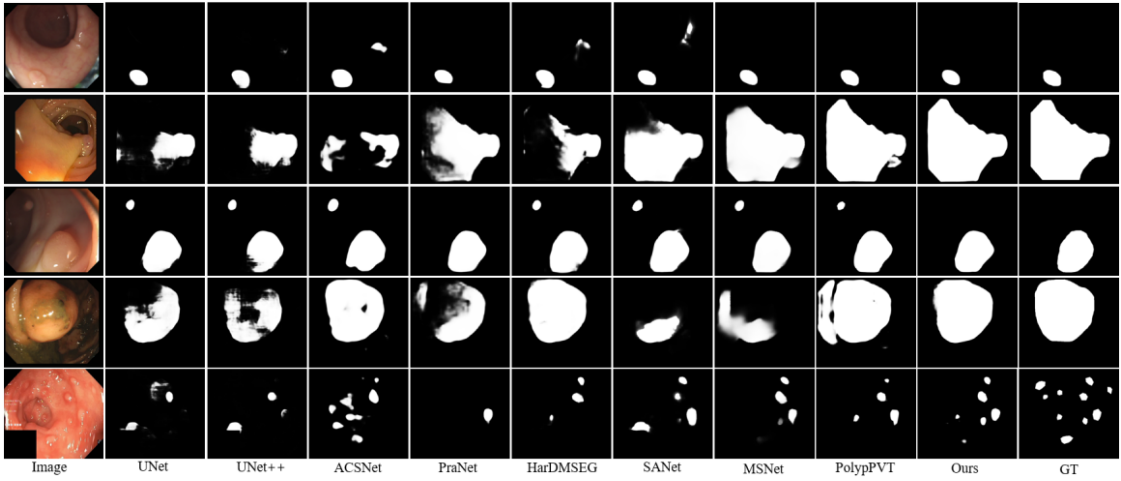

**Fig. 5.** Visualization comparison with state-of-the-art methods

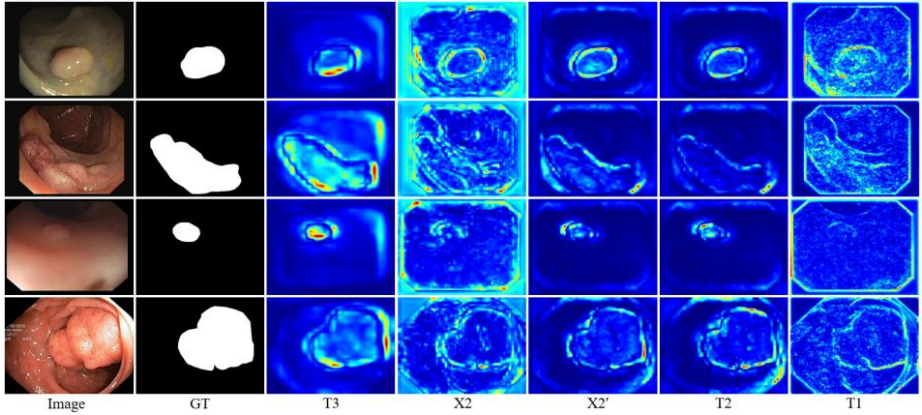

**Fig. 6.** Visualization of feature maps. GT denotes Ground Truth. $T_2, T_3$ represent the outputs of HFEM. $X_2$ stands for the output of the second layer of the encoder. $X_2'$ is the feature map of $X_2$ after the aggregation part in HFEM. $T_1$ is the output of MAM.

*5.1. Comparison With the State-of-the-Art Methods*

To verify the effectiveness of MLFF-Net, we compare the proposed method with several state-of-the-art methods, including UNet[18], UNet++[66], SFA[67], PraNet[51], ACSNet[24], MSEG[68], MSNet[69], EUNet[25], SANet[26], FAPN[70], Polpy-PVT[29], and SAM[71]. The comparison results are shown in Fig.5 and Table 1.

Fig.5 shows the partial polyp segmentation results predicted by MLFF-Net with competing networks. Compared to other methods, MLFF-Net can predict various polyps. The tiny polyps in the first row are highly similar to the background tissue in terms of color, and texture. A part of the methods suffer from over-segmentation, such as UNet++ and ACSNet misclassifying the adherent marginal tissue near the polyp portion as a polyp and not accurately identifying the polyp edge. In contrast, the segmentation results of our method are smoother and closer to the GT. The polyps in the second row differed from regular polyps in size and shape. Regular polyps tend to show rounded bumps, while the second-row polyp shape exhibits a flattened shape and larger size. Unet++ and HarDMSEG only identify the anterior part of the polyp. MSNet and Polyp-PVT, although predict the entire polyp tissue, show over-segmentation in the anterior edge part. Our model can accurately predict the overall polyps. In the third row, a normal tissue protrusion is

*\* Corresponding author.*
E-mall addresses: chenqs@cqupt.edu.cn (Q. Chen)

present in the upper left portion of the image. Most methods misclassify tissue protrusions as polyps. In comparison, MLFF-Net can accurately locate the polyp and eliminates mistaken identity. The challenge in segmenting polyps in the fourth row is the presence of tissue mucus on the polyp, which varies widely in color and texture from the polyp. Therefore, most methods produce discontinuous predictions. For example, PraNet and SANet misidentified the tissue mucus as the polyp boundary and identified only a portion of the polyp. ACSNet and Polyp-PVT produced holes in the prediction results. For the case of multiple polyps in a single colon image, as shown in the fifth row, most methods can only identify a few polyps (e.g., PraNet, HarDMSEG) and are prone to false predictions (e.g., ACSNet, SANet). Although MLFF-Net is not able to accurately segment all polyps, it has a greater improvement in the number and

**Table 1**

Quantitative results and comparison on CVC-ClinicDB, Kvasir.

| Dataset | Model | $mDic$ | $mIou$ | $F_\beta^\omega$ | $S_\alpha$ | $mE_\xi$ | $maxE_\xi$ | $MAE$ |
|---|---|---|---|---|---|---|---|---|
| CVC-ClinicDB | UNet[18] | 0.823 | 0.755 | 0.811 | 0.889 | 0.913 | 0.954 | 0.019 |
| | UNet++[66] | 0.794 | 0.729 | 0.785 | 0.873 | 0.891 | 0.931 | 0.022 |
| | SFA[67] | 0.700 | 0.607 | 0.647 | 0.793 | 0.840 | 0.885 | 0.042 |
| | ParNet[51] | 0.899 | 0.849 | 0.896 | 0.936 | 0.963 | 0.979 | 0.009 |
| | ACSNet[24] | 0.882 | 0.826 | 0.873 | 0.927 | 0.947 | 0.959 | 0.011 |
| | MSEG[68] | 0.909 | 0.864 | 0.907 | 0.938 | 0.961 | 0.969 | 0.007 |
| | MSNet[69] | 0.921 | 0.879 | 0.914 | 0.941 | ~ | 0.972 | 0.008 |
| | EUNet[25] | 0.902 | 0.846 | 0.891 | 0.936 | 0.959 | 0.965 | 0.011 |
| | SANet[26] | 0.916 | 0.859 | 0.909 | 0.939 | 0.971 | 0.976 | 0.012 |
| | FAPN-s[70] | 0.931 | 0.879 | 0.929 | 0.941 | ~ | ~ | 0.008 |
| | Polpy-PVT[29] | 0.937 | 0.889 | 0.936 | 0.949 | 0.985 | 0.989 | **0.006** |
| | SAM-L[71] | 0.579 | 0.526 | 0.563 | 0.744 | ~ | 0.685 | 0.057 |
| | Ours | **0.943** | **0.897** | **0.944** | **0.952** | **0.988** | **0.992** | **0.006** |
| Kvasir | UNet[18] | 0.818 | 0.746 | 0.794 | 0.858 | 0.881 | 0.893 | 0.055 |
| | UNet++[66] | 0.821 | 0.743 | 0.808 | 0.862 | 0.886 | 0.909 | 0.048 |
| | SFA[67] | 0.723 | 0.611 | 0.670 | 0.782 | 0.834 | 0.849 | 0.075 |
| | ParNet[51] | 0.898 | 0.840 | 0.885 | 0.915 | 0.944 | 0.948 | 0.030 |
| | ACSNet[24] | 0.898 | 0.838 | 0.882 | 0.920 | 0.941 | 0.952 | 0.032 |
| | MSEG[68] | 0.897 | 0.839 | 0.885 | 0.912 | 0.942 | 0.948 | 0.028 |
| | MSNet[69] | 0.907 | 0.862 | 0.893 | 0.922 | ~ | 0.944 | 0.028 |
| | EUNet[25] | 0.908 | 0.854 | 0.893 | 0.917 | 0.951 | 0.954 | 0.028 |
| | SANet[26] | 0.904 | 0.847 | 0.892 | 0.915 | 0.949 | 0.953 | 0.028 |
| | FAPN-s[70] | 0.908 | 0.858 | 0.898 | 0.913 | ~ | ~ | 0.028 |
| | Polpy-PVT[29] | 0.917 | 0.864 | **0.911** | 0.925 | 0.956 | **0.962** | **0.023** |
| | SAM-L[71] | 0.782 | 0.710 | 0.773 | 0.832 | ~ | 0.836 | 0.061 |
| | Ours | **0.919** | **0.866** | 0.908 | **0.925** | **0.958** | **0.962** | **0.023** |

**Table 2**

Quantitative results and comparison on CVC-ColonDB, CVC-300, ETIS.

| Dataset | Model | $mDic$ | $mIou$ | $F_\beta^\omega$ | $S_\alpha$ | $mE_\xi$ | $maxE_\xi$ | $MAE$ |
|---|---|---|---|---|---|---|---|---|
| CVC-ColonDB | UNet[18] | 0.512 | 0.444 | 0.498 | 0.712 | 0.696 | 0.776 | 0.061 |
| | UNet++[66] | 0.483 | 0.410 | 0.467 | 0.691 | 0.680 | 0.760 | 0.064 |
| | SFA[67] | 0.469 | 0.347 | 0.379 | 0.634 | 0.675 | 0.764 | 0.094 |
| | ParNet[51] | 0.712 | 0.640 | 0.699 | 0.820 | 0.847 | 0.872 | 0.043 |
| | ACSNet[24] | 0.716 | 0.649 | 0.697 | 0.829 | 0.839 | 0.851 | 0.039 |
| | MSEG[68] | 0.735 | 0.666 | 0.724 | 0.834 | 0.859 | 0.875 | 0.038 |
| | MSNet[69] | 0.755 | 0.678 | 0.737 | 0.836 | ~ | 0.883 | 0.041 |
| | EUNet[25] | 0.756 | 0.681 | 0.730 | 0.831 | 0.863 | 0.872 | 0.045 |
| | SANet[26] | 0.753 | 0.670 | 0.726 | 0.837 | 0.869 | 0.878 | 0.043 |
| | FAPN-s[70] | 0.780 | 0.702 | 0.768 | 0.846 | ~ | ~ | 0.034 |
| | Polpy-PVT[29] | 0.808 | 0.727 | 0.795 | 0.865 | 0.913 | **0.919** | **0.031** |
| | SAM-L[71] | 0.468 | 0.422 | 0.463 | 0.690 | ~ | 0.608 | 0.054 |
| | Ours | **0.820** | **0.742** | **0.805** | **0.870** | **0.914** | 0.917 | 0.035 |
| CVC-300 | UNet[18] | 0.710 | 0.627 | 0.684 | 0.843 | 0.847 | 0.875 | 0.022 |
|  | UNet++[66] | 0.707 | 0.624 | 0.687 | 0.839 | 0.834 | 0.898 | 0.018 |
|  | SFA[67] | 0.467 | 0.329 | 0.341 | 0.640 | 0.644 | 0.817 | 0.065 |
|  | ParNet[51] | 0.871 | 0.797 | 0.843 | 0.925 | 0.950 | 0.972 | 0.010 |
|  | ACSNet[24] | 0.863 | 0.787 | 0.825 | 0.923 | 0.939 | 0.968 | 0.013 |
|  | MSEG[68] | 0.874 | 0.804 | 0.852 | 0.924 | 0.948 | 0.957 | 0.009 |
|  | MSNet[69] | 0.869 | 0.807 | 0.849 | 0.925 | ~ | 0.943 | 0.010 |
|  | EUNet[25] | 0.837 | 0.765 | 0.805 | 0.904 | 0.919 | 0.933 | 0.015 |
|  | SANet[26] | 0.888 | 0.815 | 0.859 | 0.928 | 0.962 | 0.972 | 0.008 |
|  | FAPN-s[70] | **0.909** | **0.845** | **0.894** | 0.937 | ~ | ~ | **0.005** |
|  | Polpy-PVT[29] | 0.900 | 0.833 | 0.884 | 0.935 | **0.973** | **0.981** | 0.007 |
|  | SAM-L[71] | 0.726 | 0.676 | 0.729 | 0.849 | ~ | 0.826 | 0.020 |
|  | Ours | 0.904 | 0.841 | 0.891 | **0.939** | 0.969 | 0.977 | 0.007 |
| ETIS | UNet[18] | 0.398 | 0.335 | 0.366 | 0.684 | 0.643 | 0.740 | 0.036 |
|  | UNet++[66] | 0.401 | 0.344 | 0.390 | 0.683 | 0.629 | 0.776 | 0.035 |
|  | SFA[67] | 0.297 | 0.217 | 0.231 | 0.557 | 0.531 | 0.632 | 0.109 |
|  | ParNet[51] | 0.628 | 0.567 | 0.600 | 0.794 | 0.808 | 0.841 | 0.031 |
|  | ACSNet[24] | 0.578 | 0.509 | 0.530 | 0.754 | 0.737 | 0.764 | 0.059 |
|  | MSEG[68] | 0.700 | 0.630 | 0.671 | 0.828 | 0.854 | 0.890 | 0.015 |
|  | MSNet[69] | 0.719 | 0.664 | 0.678 | 0.840 | ~ | 0.830 | 0.020 |
|  | EUNet[25] | 0.687 | 0.609 | 0.636 | 0.793 | 0.807 | 0.841 | 0.067 |
|  | SANet[26] | 0.750 | 0.654 | 0.685 | 0.849 | 0.881 | 0.897 | 0.015 |
|  | FAPN-s[70] | 0.750 | 0.674 | 0.726 | 0.849 | ~ | ~ | **0.012** |
|  | Polpy-PVT[29] | **0.787** | 0.706 | **0.750** | **0.871** | **0.906** | **0.910** | 0.013 |
|  | SAM-L[71] | 0.551 | 0.507 | 0.544 | 0.751 | ~ | 0.687 | 0.030 |
|  | Ours | 0.784 | **0.707** | **0.750** | 0.866 | 0.891 | 0.903 | 0.025 |

accuracy of recognition.

To fully understand how the model works, the deep feature maps in the network are visualized and the results are shown in Fig.6. Take the results displayed in the first row as an example. $T_3$ (shown in the third column), which aggregates semantic information in HFEM, has the ability to fuzzily localize polyps. $X_2$ is the output of the second layer encoder and contains a lot of noise that is harmful to segmentation. $X_2'$ is generated by fusing $X_2$ with the deeper feature maps. Compared with $X_2$, $X_2'$ removes most of the useless information and focuses only on the polyp part, as shown in the fifth column. $T_2$ is the feature map after $X_2'$ has gone through the attention module in HFEM. It corrects the conflicts and redundancies that might generate during feature fusion. Information that does not belong to the polyp appears in the upper left part of the polyp in $X_2'$, and this error is fixed in $T_2$. The feature map $T_1$ from MAM contains a large amount of detailed information and tiny tissues. It is worth noting that for the shallow output $T_1$, our model pays more attention to the contour details of the polyps.

Table 1 shows the results of MLFF-Net on CVC-ClinicDB and Kvasir, which reflect the learning ability of the model. MLFF-Net has $mDic$ of 0.943, $mIoU$ of 0.897, weighted F-measure of 0.944, S-measure of 0.952, mean E-measure of 0.988, max E-measure of 0.992, and $MAE$ of 0.006 on CVC-ClinicDB. Compared with current advanced polyp segmentation methods, MLFF-Net achieved the best performance on all evaluation criteria. On Kvasir, our method outperforms most existing studies and is comparable to the best method.

The results on CVC-ColonDB, CVC-300, and ETIS, which are illustrated in Table 2, reflect the generalization ability of our model. The above datasets are not involved in the training and differ significantly from the training data in terms of color, and background. On CVC-ColonDB, our method improves 1.5%, 2%, 1.3%, 0.6%, and 0.1% over one of the best methods (polyp-PVT) for $mDic$, $mIoU$, weighted F-measure, S-measure, and mean E-measure, respectively. On the CVC-300 and ETIS, our method does not reach optimal results on all metrics but still outperforms most studies in the same field. In general, MLFF-Net has good performance in both learning ability and generalization ability, as well as demonstrates excellent and balanced performance on multiple datasets.

*5.2. Ablation Experiment*

The effect of each module is discussed through ablation experiments. To be fair, all experiments are set up consistently except for changes in the model architecture. The UNet of the PVTv2 encoder is used as the baseline (Bas.). MAM, HFEM, and GAM are gradually added to the baseline network, and $mDic$ and $mIoU$ are calculated. The results of the ablation experiments are shown in Table 3, which includes $mDic$ and $mIoU$ values on five datasets. MAM captures multi-scale information and detailed information in shallow features. The network with MAM added improves $mDic$ by 1.8%, 0.4%, 3%, and 1.7%, and $mIoU$ by 2%, 0.7%, 3%, and 2.4% on CVC-ClinicDB, Kvasir, CVC-ColonDB, and ETIS compared to the baseline. Although $mDic$ and

*mIoU* are reduced on the CVC-300 dataset, the overall performance is more balanced across the five datasets. HFEM is applied to aggregate deep information and mitigate aggregation conflicts. The addition of HFEM allows the network to perform better on multiple datasets. Finally, the addition of GAM constitutes the proposed overall framework. This suggests that our network can accurately capture the structural features of polyps with less influence by color, light differences due to inconsistent image collection conditions.

Fig.7 shows the segmentation results of different models in the ablation experiment. The Bas. may show incorrect predictions. For example, in the blue box in the first row, Bas. predicts the flat tissue around the polyp as the lesion tissue. In the second row of the red box, Bas. mispredicts the shiny location in the center of the polyp as an intact polyp. These tissues differ greatly from polyps and do not have polyp bulge-like features. The addition of MAM preserves the detail and multi-scale information of the polyps. As shown in the red boxes in the first and

**Table 3**

Results of ablation experiments on 5 datasets second rows, MAM corrects the part of the polyp not predicted by Bas. and makes the prediction more complete. However, the drawback of MAM is that it tends to capture some parts that are highly similar to the polyp tissue, such as the circular bump in the green box in the first row, the flat bump in the blue box in the second row, and the tissue fluid near the polyp in the red box in the third row. HFEM enhances the network's ability to localize and correct polyp locations. It fixes the localization error that occurs in the first and second rows of results, and furthermore accurately predicts the margins of polyps. Unfortunately, unavoidable errors may occur during feature fusion, such as the incorrect prediction of the green box in the second row. GAM works on the decoder, which can affect the entire upsampling and prediction process of the feature map. As shown in the fifth column, it fixes the erroneous prediction present in the above network and the prediction results are closest to GT.

|  | CVC-ClinicDB | | Kvasir | | CVC-ColonDB | | CVC-300 | | ETIS | |
| --- | --- | --- | --- | --- | --- | --- | --- | --- | --- | --- |
|  | mDice | mIoU | mDice | mIoU | mDice | mIoU | mDice | mIoU | mDice | mIoU |
| Bas. | 0.912 | 0.864 | 0.908 | 0.854 | 0.787 | 0.710 | 0.899 | 0.830 | 0.759 | 0.679 |
| +MAM | 0.928 | 0.881 | 0.912 | 0.860 | 0.811 | 0.731 | 0.886 | 0.815 | 0.772 | 0.695 |
| +MAM +HFEM | 0.938 | 0.891 | 0.920 | 0.872 | 0.804 | 0.725 | 0.895 | 0.827 | 0.779 | 0.700 |
| Ours | 0.943 | 0.897 | 0.919 | 0.866 | 0.820 | 0.742 | 0.904 | 0.841 | 0.784 | 0.707 |

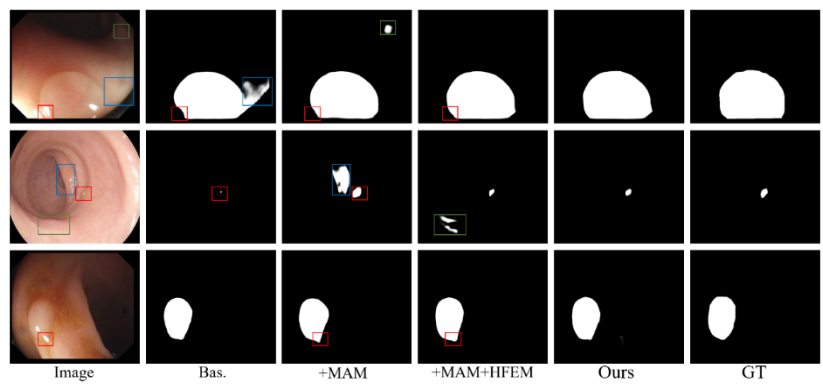

**Fig. 2.** Visualization of the ablation study results. The red, blue, and green boxes represent the locations where the prediction deviation occurred.

*5.3. Limitations*

Although MLFF-Net has a more balanced overall performance on multiple datasets and can accurately segment multiple types of polyps compared to other superior methods. Nonetheless, it still has limitations. On the one hand, the data in this study are based on image frames abstracted from colonoscopy videos. In the clinical setting, automatic polyp segmentation algorithms based on colonoscopy videos are more urgently required. Unfortunately, current video-based algorithms are generally less effective and require more computational resources due to limitations in the dataset and algorithm complexity. On the other hand, the prediction results of MLFF-Net are not perfect for samples with multiple polyps in the image. This is due to two reasons: 1) there are too little training data for multiple polyps in the same image, and 2) these polyps are closer to normal protruding tissues, making segmentation more challenging. In the future, we will propose more effective automatic segmentation algorithms for this polyp type.

## 6. Conclusion

In this paper, we propose a Multi-level feature fusion network combining attention mechanisms for polyp segmentation, called MLFF-Net. Firstly, MAM is suggested for extracting multiscale information and detailed information from shallow features. Secondly, HFEM is used to aggregate deep semantic information and alleviate the semantic conflict problem in the fusion process. Finally, GAM is designed to solve the decoder local receptive field problem. It not only captures global relations but also fuses same-level encoder and decoder features. Experiments on Kvasir, CVC-ClinicDB, CVC-ColonDB, ETIS, and CVC-300 and visualization of feature maps show that MLFF-Net not only has superior learning ability and generalization ability, but also has the capability to accurately segment multiple types of polyps.